\documentclass[conference]{IEEEtran}
\usepackage{hyperref}
\usepackage{ifpdf}
\usepackage{graphicx}
\usepackage{subcaption}
\usepackage{amsmath}
\usepackage{ amssymb }
\usepackage{subcaption}
\usepackage[rightcaption]{sidecap}
\usepackage{xcolor}
\usepackage[linesnumbered,ruled,vlined]{algorithm2e}
\usepackage{algorithmicx}
\usepackage{multirow}

\graphicspath{ {./} }

\begin{document}


\title{A Hybrid CNN-LSTM model for Video Deepfake Detection by Leveraging Optical Flow Features}

\author{\IEEEauthorblockN{Anonymous Authors}}





\author{\IEEEauthorblockN{Pallabi Saikia\textsuperscript{1,*}, Dhwani Dholaria\textsuperscript{1}, Priyanka Yadav\textsuperscript{1}, Vaidehi Patel\textsuperscript{1}, Mohendra Roy\textsuperscript{2,*}}
\IEEEauthorblockA{\textit{\textsuperscript{1}Computer Science and Engineering Department, School of Technology} \\ \textit{\textsuperscript{2}Information \& Communication Technology, School of Technology} \\ \textit{Pandit Deendayal Energy University, Gandhinagar -382007, India} \mbox{}\\  \textit{*Corresponding Authors: mohendra.roy@ieee.org; pallabi.iitg@gmail.com}  }}


\maketitle

\begin{abstract}

Deepfakes are the synthesized digital media in order to create ultra-realistic fake videos to trick the spectator.  Deep generative algorithms, such as, Generative Adversarial Networks( GAN) are widely used to accomplish such tasks. This approach synthesizes pseudo-realistic contents that are very difficult to distinguish by traditional  detection methods. In most cases, Convolutional Neural Network(CNN) based discriminators are being used for detecting such synthesized media. However, it emphasise primarily on the spatial attributes of individual video frames, thereby fail to learn the temporal information from their inter-frame relations. In this paper, we leveraged an optical flow based feature extraction approach to extract the temporal features, which are then fed to a hybrid model for classification. This hybrid model is based on the combination of CNN and recurrent neural network (RNN) architectures. The hybrid model provides effective performance on open source data-sets such as, DFDC, FF++ and Celeb-DF. This proposed method shows an accuracy of $66.26\%$, $91.21\%$ and $79.49\%$ in DFDC, FF++ , and Celeb-DF respectively with a very reduced No of sample size of $\le$ 100 samples(frames). This promises early detection of fake contents compared to existing modalities.


\end{abstract}

\footnote{\copyright IEEE, Paper No: 832, IJCNN, 2022 IEEE World Congress on Computational Intelligence }

\section{Introduction}

Deepfakes are images and videos, usually created by deep neural networks to superimpose target subject's face features over another in order to produce fake media. According to a recent report from the Deeptrace Lab \cite{kikerpill2020choose}, there are almost 15,000 deefake media over the internet consisting of non-obscene videos targeting the politicians and the functioning of democratic societies. More than 13,000 deepfake videos were found on different deepfake-specific porn sites \cite{ajder2019state}, and about 96\% of the deepfake content on web are related to pornography and mostly related to famous celebrities \cite{ajder2019state} to defame the individuals. Day by day, deepfake's findings are becoming so realistic that they are almost indistinguishable, and the substituted subjects are rigged to say things they never spoke \cite{tolosana2020deepfakes}. Deepfake methods have been extensively applied nowadays to produce enormous fake news, posing a serious threat to communities worldwide, and have the potential to influence the masses as well as democratic and geopolitical structure of a region \cite{chesney2019deep}.


Many organizations as well as private companies are investing heavily to counter the challenges of deepfake. Various techniques have already been studied in the field of deepfake detection, including Machine Learning (ML) techniques such as Support Vector Machines (SVMs) \cite{kharbat2019image}, and Deep learning techniques such as Convolution Neural Network(CNN) \cite{afchar2018mesonet}, CNN with SVM \cite{yang2019exposing}, Recurrent Neural Network(RNN) \cite{guera2018deepfake}, CNN with Long Short Term Memory (LSTM) \cite{guera2018deepfake}, etc. Moreover, many traditional ways have also been explored to detect manipulated media, such as exposing inconsistent in head-poses, consideration of the background color manipulation \cite{agarwal2020detecting}. However, most of these techniques are focuses primarily on the spatial feature analysis and does not include any temporal information. Since, most of the deepfake media are in the form of video, therefore, identifying inconsistencies in temporal information (along with spatial inconsistencies) may enhance the classification accuracy of deepfakes. 

In this work, we investigated a hybrid deep learning approach on modelling the intra-frame as well as inter-frame features of videos to accurately identify its authenticity. Further, we incorporated a traditional temporal feature analysis method, optical flow to help in extraction of the temporal features. The optical flow implementation is based on to characterization of the motion of the subject’s face and the technique exploits the possible inter-frame dissimilarities. A detail analysis has been carried out on to characterize the proposed model on various sets of video data. The proposed model is evaluated based on its various performance parameters such as Accuracy, Recall, Precision, F1-score, and AUC.

\section{Background and Related Works}

\subsection{Deepfake Generation}


There are several machine learning algorithms that can produce credible deceptive videos. Moreover, the recent advancement in adversarial techniques such as generative adversarial network (GAN) has fuel the rapid development of digital forgery. The algorithms have been widely used in many of the modern deepfake generation approaches\cite{karras2020analyzing, singh2020using}. The use of Generative Adversarial Nets (GANs) in deepfake generation has been a prominent method based on neural networks \cite{goodfellow2014generative}. It works on the idea of setting dual neural networks in conflict with one another, i.e., the generator G that generates the output image and the discriminator D that determines whether its fake or real \cite{singh2020using}. GAN was first introduced in 2014 by Goodfellow et al. \cite{goodfellow2014generative}. The generator G generates fake data $x_{g}$ to mislead the discriminator $D$. $D$ also learns how to differentiate between the fake media ($x_{g} = G\left(x\right)$ where $z\sim N$) and real media ($x\in X$). $G$ and $D$ are trained on an adversarial loss respectively as follows,
 \begin{equation}
 L_{adv}\left(D\right) = \max\log D\left(x\right) + \log \left(1- D\left(G\left(z\right)\right)\right)
 \end{equation}
 \begin{equation}
 L_{adv}\left(G\right) =\min \log \left(1-D\left(G\left(z\right)\right)\right)
 \end{equation}

There are two major approaches in deepfake generation, these are FaceSwap and Face Synthesis. In the FaceSwap, a target face is swapped onto a source face and in Face synthesis the facial features are being synthesised. Recent High-Resolution Face Swapping method from Disney Research is one of the very successful face swapping method \cite{naruniec2020high}. Similarly the LandmarkGAN is a Face Synthesis method based on facial landmark as input \cite{sun2020landmarkgan}. All these state-of-art techniques are capable of generating ultra-realistic synthesized media, which are almost indistinguishable by traditional means. This demands a sophisticated detection technique. Recently, the adversarial detectors are being used to counter this issue. However, in many cases the adversarial detectors can be tricked \cite{akhtar2021advances}. Therefore, a more generalized multi-model approach is focused in nowadays research.



\subsection{Deepfake Detection}


Traditionally, inconsistencies or unrealistic elements in the forgeries are targeted by several detecting approaches \cite{nguyen2019deep}. Most of the detection techniques nowadays, mainly rely on machine learning techniques to generalise the detection process \cite{maksutov2020methods}. Peng Chen, et. al.  \cite{chen2020fsspotter} has developed FSSPOTTER, a unified framework for detection of deepfakes. It investigates the rich spatial features within a single frame with the help of a Spatial Feature Extractor (SFE) along with a Temporal Feature Aggregator (TFA) which extracts the inconsistencies between the frames. Digvijay Yadav, et al. \cite{yadav2019deepfake} considered blinking of eyes as one of the important features to detect the deepfake, and for detecting the DeepFakes, CNN architecture is combined with LSTM to detect the temporal inconsistencies in changes in the frames. Irene Amerini, et. al. \cite{amerini2019deepfake} introduced a method to exploit the temporal inconsistencies of videos with optical flow fields between two consecutive frames so as to differentiate between original and fake videos. Shivangi et al. \cite{aneja2020generalized} proposed a transfer learning based approach, named as Deep Distribution Transfer, to overcome the problem of zero-shot and few-shot transfer for facial forgery detection. Distribution-based loss formulation is used to bridge the gaps between different facial forgery techniques involved in the creations of the datasets. XTao, et. al. \cite{tao2017detail} proposed another framework that accentuates the way to accomplish better outcomes, appropriate edge arrangement and motion compensation. The work consists of introducing a ``sub-pixel motion compensation” (SPMC) layer in a CNN framework. The versatile CNN structure joins the SPMC layer and fuses numerous frames to reveal image details. The paper furnish an investigation into how to coordinate numerous casing inputs for improving outcomes. David Guera, et. al. \cite{guera2018deepfake} demonstrated the effectiveness of Long Short-Term Memory (LSTM) networks along with CNN model for the detection of the deepfakes. InceptionV3 with fully connected layer at the top of the network outputs a deep representation of each frame and further LSTM model takes the feature sequences to model the temporal dependencies. This method thus elaborates a temporal-aware system to automatically detect deepfake videos. Typically, well-known and pre-trained CNN algorithms are widely explored in literature to learn discrete aspects from each frame of the video sequence. Most of the state-of-the-art algorithms focus primarily on extraction of the intra-frame features for deeepfake detection. However, effective extraction of inter-frame feature to exploit temporal inconsistencies is also a promising direction in the research of deepfake detection.

\begin{figure*}[!htb]
\centering
\includegraphics[width=0.99\linewidth]{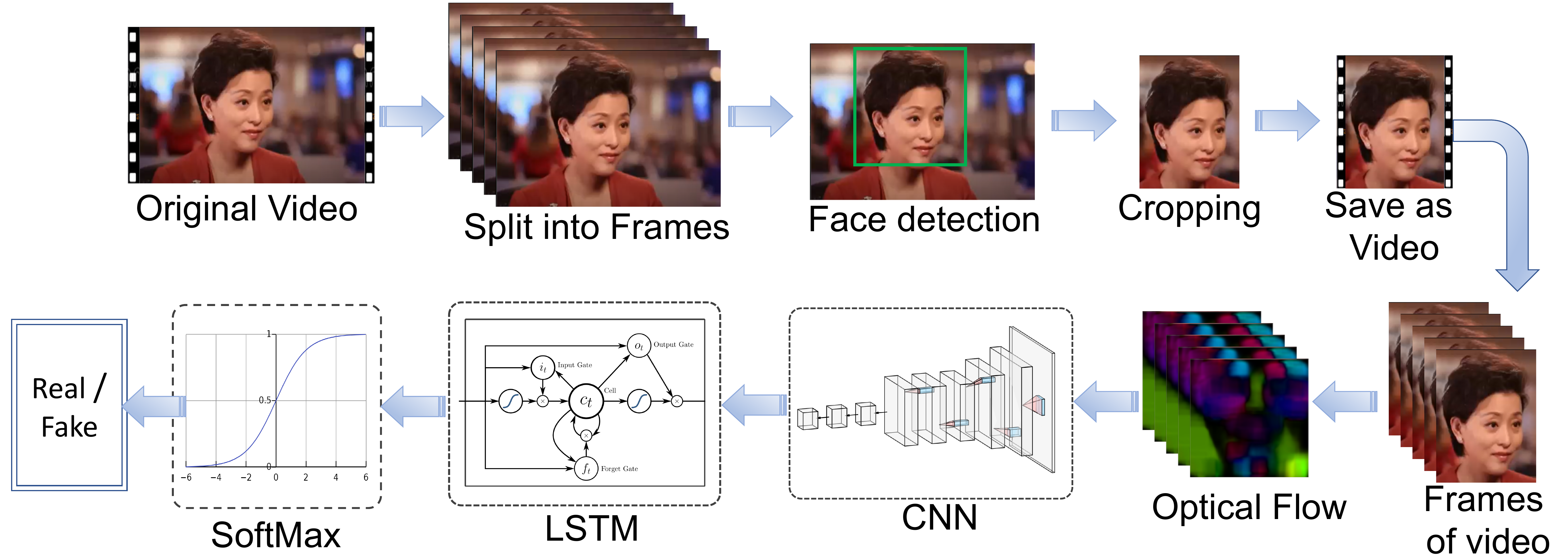}
\caption{Proposed Workflow of Deepfake Detection in video. Here, the original video was splitted into frames and then the face ROI was segmented out by cropping the ROI. The cropped frames are again saved as video. This reduced sizes videos are then used for generating optical flow. The optical flow features are then fed as input feature to the hybird classification model consisting of combination of CNN and LSTM. The output of the LSTM were finally activated through SoftMax function to find the probability of classes. Thus obtain the classification.}
\label{fig:figname}
\centering
\end{figure*}

\section{Proposed Method}

In this work, we have focused on the facial features of a video to find characteristics of a deepfake content. The most common characteristics in the manipulated media is that an individual in a video is supplanted with another person's profile. On account of that we are contemplating the facial characteristics, as the warping leaves a few distinguishable artifacts in the deepfake videos. We exploit this information and incorporated various pre-processing steps, followed by model building as illustrated below: 




\subsection{Frame Extraction}

To grab the frames from the video, uniform sampling is applied on the video duration. During preprocessing, we disregarded the frames that doesn't contain a face. Handling highly sampled frames will require a great deal of computational power. So, for probing reason on an average we have extracted 148 frames per video. Further, we have extracted the face of a subject from these videos as mentioned below.




\subsection{Face Extraction}
The proposed methodology considers the face as a region of interest (ROI). We extracted the ROI using a batch\_face\_locations algorithm within face\_recognition library
\cite{geitgey2019face}. Face\_Recognition library wraps dlib's face recognition functions \cite{king2009dlib} into a simple, easy to use API. It captures 128 data points per face, resulting in unique parameters for the hash. A re-scaling was performed to remove extra-remaining background to reduce the memory complexity, which eventually help in reducing the computational complexity of the model. Thus, we obtain a modified video data-set with frames having size of $112\times112.$

\subsection{Optical flow field Feature Extraction}

Optical flow fields for consecutive pairs of frames are generated to detect the pattern of apparent motion in the individual pixels on the image plane \cite{horn1981determining}. It is used to extricate the movement of patterns in an image based on apparent velocities distribution. The image intensity between consecutive frames can be expressed as a function of time ($t$) and space ($x$,$y$). Thus, an frame or image can be represented as $I (x, y, t)$. If the image makes a displacement of ($dx$, $dy$) in time $dt$, then the new image will be $I(x+dx, y+dy, t+dt)$.





Using the Taylor series expansion, the change can be expanded as: 

\vspace{-2mm}

\begin{equation}
    I(x+dx, y+dy, t+dt) = I(x, y, t) + \frac{\partial{I}}{\partial{x}}dx + \frac{\partial{I}}{\partial{y}}dy + \frac{\partial{I}}{\partial{t}}dt +...
\end{equation}

If the change in intensities remain constant in both the frames, then $I(x,y,t) = I(x+dx, y+dy, t+dt)$


\begin{equation}\label{eq:diff}
    \Rightarrow \frac{\partial{I}}{\partial{x}}dx + \frac{\partial{I}}{\partial{y}}dy + \frac{\partial{I}}{\partial{t}}dt = 0
\end{equation}

		
		
Now, dividing the equation (\ref{eq:diff}) by $dt$, we get:

\begin{equation}
    \Rightarrow \frac{\partial{I}}{\partial{x}}u + \frac{\partial{I}}{\partial{y}}v + \frac{\partial{I}}{\partial{t}} = 0
\end{equation}


where, $u=\frac{dx}{dt}$ and 
$v=\frac{dy}{dt}$ \\

\hspace{5mm}
$\frac{\partial{I}}{\partial{x}} =$ image gradient along the x-axis \\

\hspace{5mm}
$\frac{\partial{I}}{\partial{y}} =$ image gradient along the y-axis \\

\hspace{5mm}
$\frac{\partial{I}}{\partial{t}} =$ image gradient along time \\
	


\begin{figure}[!h]
\centering
\includegraphics[width=1.0\linewidth]{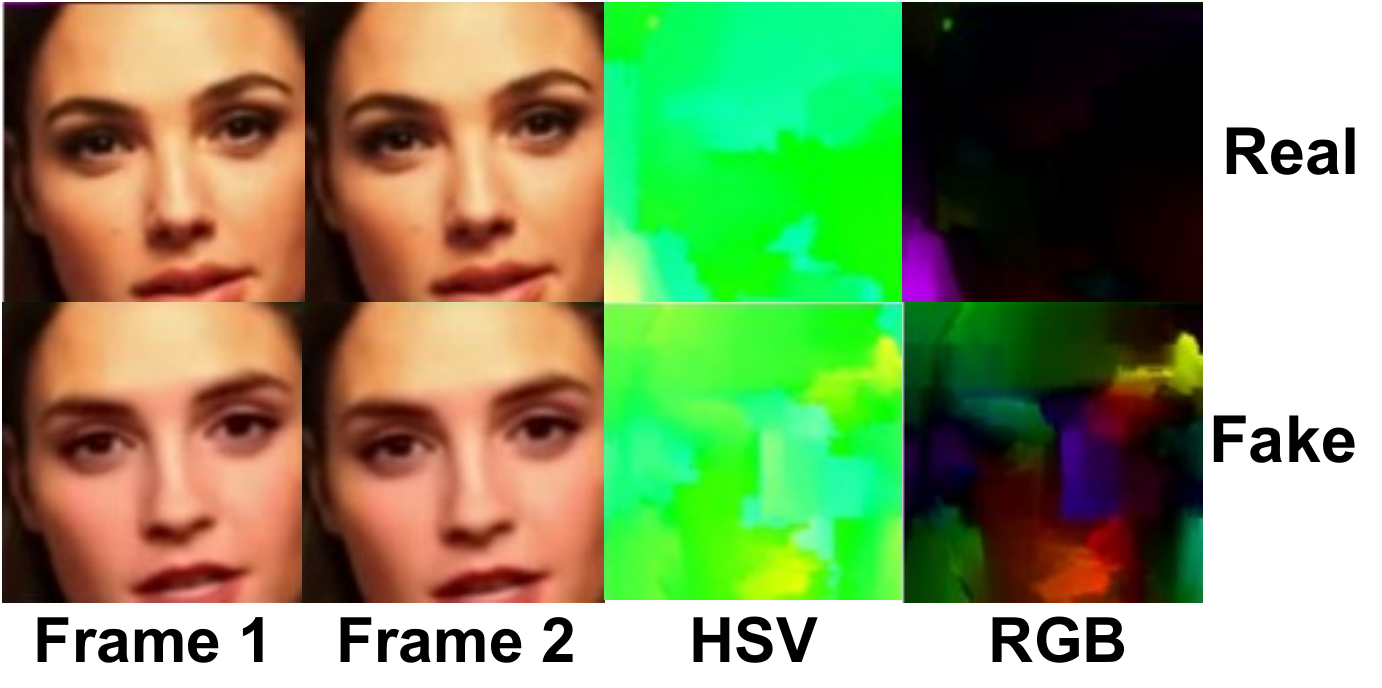}
\caption{Optical flow features of real and fake video frames. From this figure a distinct variation in the optical flow features can be visible between real and fake frames.}
\label{fig:optical_flow}
\centering
\end{figure}

The cropped face videos, that obtained in the previous step of face extraction are splitted into frames and optical flow between two consecutive frames is calculated for identifying temporal inconsistencies for deepfake detection. Through calculation of optical flow, we detected the change in motion of every pixel of an image. The HSV color representation of the optical flow vector using the magnitude and the hue component envisioned by the direction vector is shown in Figure \ref{fig:optical_flow}. From our study, we find that the fake videos have distorted motion vectors as compared to the real ones. This motion vector plot of each axis was then converted to 3 channel images using predefined color code for feeding into the hybrid model discussed in the next step.

\subsection{Hybrid CNN-RNN Architecture for modelling}

\begin{table*}[]
\centering
\caption{Comparison of the performance of the various base models}
\label{tab:table6}
\resizebox{\linewidth}{!}{%
\begin{tabular}{|c|ccc|ccc|ccc|}
\hline
 &
  \multicolumn{3}{c|}{\textbf{DFDC (Frames:20)}} &
  \multicolumn{3}{c|}{\textbf{Celeb-DF (Frames:50)}} &
  \multicolumn{3}{c|}{\textbf{FF++ (Frames:30)}} \\ \hline
\textbf{Batch Size:128} &
  \multicolumn{1}{c|}{\textbf{Test Accuracy}} &
  \multicolumn{1}{c|}{\textbf{F1}} &
  \textbf{AUC} &
  \multicolumn{1}{c|}{\textbf{Test Accuracy}} &
  \multicolumn{1}{c|}{\textbf{F1}} &
  \textbf{AUC} &
  \multicolumn{1}{c|}{\textbf{Test Accuracy}} &
  \multicolumn{1}{c|}{\textbf{F1}} &
  \textbf{AUC} \\ \hline
\textbf{VGG16} &
  \multicolumn{1}{c|}{\textbf{64.73\%}} &
  \multicolumn{1}{c|}{\textbf{64.28\%}} &
  \textbf{0.64} &
  \multicolumn{1}{c|}{67.09\%} &
  \multicolumn{1}{c|}{68.14\%} &
  0.68 &
  \multicolumn{1}{c|}{78.39\%} &
  \multicolumn{1}{c|}{78.45\%} &
  0.78 \\ \hline
\textbf{InceptionV3} &
  \multicolumn{1}{c|}{45.19\%} &
  \multicolumn{1}{c|}{50.00\%} &
  0.5 &
  \multicolumn{1}{c|}{55.12\%} &
  \multicolumn{1}{c|}{52.27\%} &
  0.52 &
  \multicolumn{1}{c|}{51.00\%} &
  \multicolumn{1}{c|}{50.51\%} &
  0.5 \\ \hline
\textbf{ResNet50} &
  \multicolumn{1}{c|}{64.58\%} &
  \multicolumn{1}{c|}{59.87\%} &
  0.59 &
  \multicolumn{1}{c|}{67.09\%} &
  \multicolumn{1}{c|}{68.44\%} &
  0.68 &
  \multicolumn{1}{c|}{\textbf{89.67\%}} &
  \multicolumn{1}{c|}{\textbf{89.65\%}} &
  \textbf{0.89} \\ \hline
\textbf{Xception} &
  \multicolumn{1}{c|}{63.20\%} &
  \multicolumn{1}{c|}{64.19\%} &
  0.64 &
  \multicolumn{1}{c|}{59.22\%} &
  \multicolumn{1}{c|}{63.49\%} &
  0.63 &
  \multicolumn{1}{c|}{73.86\%} &
  \multicolumn{1}{c|}{73.04\%} &
  0.73 \\ \hline
\textbf{MobileNetV2} &
  \multicolumn{1}{c|}{57.09\%} &
  \multicolumn{1}{c|}{55.68\%} &
  0.55 &
  \multicolumn{1}{c|}{63.24\%} &
  \multicolumn{1}{c|}{65.21\%} &
  0.65 &
  \multicolumn{1}{c|}{76.63\%} &
  \multicolumn{1}{c|}{76.76\%} &
  0.76 \\ \hline
\textbf{EfficientNetB7} &
  \multicolumn{1}{c|}{59.84\%} &
  \multicolumn{1}{c|}{56.16\%} &
  0.56 &
  \multicolumn{1}{c|}{\textbf{70.08\%}} &
  \multicolumn{1}{c|}{\textbf{69.13\%}} &
  \textbf{0.69} &
  \multicolumn{1}{c|}{83.66\%} &
  \multicolumn{1}{c|}{84.04\%} &
  0.84 \\ \hline
\end{tabular}%
}
\end{table*}

\begin{table*}[!h]
\centering
\caption{Comparison of the performance of the base models with optical flow as input features on various datasets}
\label{tab:table4}
\fontsize{20}{25}\selectfont
\resizebox{\textwidth}{!}{%
\begin{tabular}{|c|cccc|cccc|cccc|}
\hline
\multirow{2}{*}{\textbf{}} &
  \multicolumn{4}{c|}{\textbf{Celeb DF}} &
  \multicolumn{4}{c|}{\textbf{DFDC}} &
  \multicolumn{4}{c|}{\textbf{FF++}} \\ \cline{2-13} 
 &
  \multicolumn{1}{c|}{\textbf{Accuracy}} &
  \multicolumn{1}{c|}{\textbf{Precision}} &
  \multicolumn{1}{c|}{\textbf{Recall}} &
  \textbf{AUC} &
  \multicolumn{1}{c|}{\textbf{Accuracy}} &
  \multicolumn{1}{c|}{\textbf{Precision}} &
  \multicolumn{1}{c|}{\textbf{Recall}} &
  \textbf{AUC} &
  \multicolumn{1}{c|}{\textbf{Accuracy}} &
  \multicolumn{1}{c|}{\textbf{Precision}} &
  \multicolumn{1}{c|}{\textbf{Recall}} &
  \textbf{AUC} \\ \hline
\textbf{OF+RNN} &
  \multicolumn{1}{c|}{52.13\%} &
  \multicolumn{1}{c|}{26.06\%} &
  \multicolumn{1}{c|}{34.21\%} &
  0.5 &
  \multicolumn{1}{c|}{54.08\%} &
  \multicolumn{1}{c|}{24.96\%} &
  \multicolumn{1}{c|}{35.08\%} &
  0.5 &
  \multicolumn{1}{c|}{47.73\%} &
  \multicolumn{1}{c|}{23.68\%} &
  \multicolumn{1}{c|}{50\%} &
  0.5 \\ \hline
\textbf{OF+CNN} &
  \multicolumn{1}{c|}{\textbf {83.33\%}} &
  \multicolumn{1}{c|}{\textbf {83.78\%}} &
  \multicolumn{1}{c|}{\textbf {83.71\%}} &
  {\textbf{0.83}} &
  \multicolumn{1}{c|}{\textbf {69.77\%}} &
  \multicolumn{1}{c|}{\textbf {69.36\%}} &
  \multicolumn{1}{c|}{\textbf {68.64\%}} &
  {\textbf {0.68}} &
  \multicolumn{1}{c|}{89.19\%} &
  \multicolumn{1}{c|}{89.51\%} &
  \multicolumn{1}{c|}{88.92\%} &
  0.88 \\ \hline
\textbf{OF+RNN+CNN} &
  \multicolumn{1}{c|}{79.49\%} &
  \multicolumn{1}{c|}{82.49\%} &
  \multicolumn{1}{c|}{79.08\%} &
  0.79 &
  \multicolumn{1}{c|}{66.26\%} &
  \multicolumn{1}{c|}{67.11\%} &
  \multicolumn{1}{c|}{65.73\%} &
  0.66 &
  \multicolumn{1}{c|}{\textbf {91.21\%}} &
  \multicolumn{1}{c|}{\textbf {91.20\%}} &
  \multicolumn{1}{c|}{\textbf {91.21\%}} &
  {\textbf {0.91}} \\ \hline

\end{tabular}%
}
\end{table*}


The combined arrays of the color coded frames obtained as a result of motion based feature extraction process as mentioned above, provide us the dataset having explicit temopal information. This data are then fed to a pre-trained CNN model. We have explored several state-of-the-art pre-trained models, such as, VGG16 \cite{qassim2018compressed}, InceptionV3  \cite{xia2017inception}, ResNet50  \cite{wen2020transfer}, Xception  \cite{chollet2017xception}, MobileNetV2  \cite{rabano2018common}, EfficientNetB7  \cite{tan2019efficientnet}. A pre-trained model, piled up with several layers of various architectural blocks to frame an exceptionally profound network, generally perform very well and require less time to re-train. Since, it explicitly trained on a million images, e.g. ''ImageNet" \cite{deng2009imagenet}, hence effective for modelling vision related problems \cite{weiss2016survey}. 


Image classification by the above pretrained models is carried out in mainly two main phases: feature extractor with the convolution layers, and discrimination with fully connected layers of CNN. The last layers of the pretrained model are fully connected, are also called as dense layers, are specific to the task, and hence was excluded for fine tuning the models on the deepfake datasets. Two LSTM layers have been added after the conv-layers and before the fully connected layers for classification. The LSTM layers also examines the interframe inconsistencies on the extracted abstract features with a dropout of 0.5. Dropout layers are significant in training complex models because they prevent the training data from being overfit. As a result, it may be possible to avoid learning of features that only appear in later samples or batches. Finally, the architecture completes with softmax layer added at the end, that compute the probabilities of the frame sequence being either fake or real. Categorical cross-entropy loss function applied to calculate the loss of the deepfake classification model, is provided in equation \ref{eq:Cat_loss}, where $\hat y_{i}$ is the predicted score of class $i$ at the softmax layer:

\begin{equation}\label{eq:Cat_loss}
    Loss= -\sum_{i=1}^{output size} y_{i} \times \log \hat y_{i}
\end{equation}

\section{Dataset description}

We have applied our proposed methodology on three sets of data, namely FaceForensics++ \cite{rossler2019faceforensics++}, Celeb-DF \cite{li2020celeb}, and the Deep Fake Detection Challenge (DFDC) dataset \cite{dolhansky2020deepfake}. We splitted the datasets into 80:20 ratio for training and testing respectively. 


\subsection{Celeb- DF}
This dataset consists of 408 real videos sourced from YouTube and synthesized 795 videos with improvements in the usual Deep-Fake generation model. Although the visual quality of the videos is low but the quality of face-swaps seems quite realistic \cite{li2020celeb}.

\subsection{DeepFakeDetectionChallenge (DFDC)}
The dataset is one of the most recent participant within the category of Deep-Fake datasets, has been compiled by Facebook AI. Sixty Six paid actors were involved in the train and test sets, and their filmed sequences were considered to generate manipulated videos internally to avoid cross-set face-swaps. Dataset comprises a total of 5214 videos out of which 78.125\% are manipulated. They achieved high quality of manipulations by choosing pairs of similar appearances, and visual quality is high as well \cite{dolhansky2020deepfake}.

\subsection{FaceForensics++}
The forensic dataset consisting of thousand original video sequences. Four automated face manipulation methods such as, Face2Face, Deepfakes, FaceSwap, and NeuralTextures have been applied to manipulate and generate the videos sequences. The data has been sourced from 977 youtube videos and all videos contain a trackable, mostly frontal face and without occlusions, which enables automated tampering methods to generate realistic forgeries \cite{rossler2019faceforensics++}.

\section{Experimental Results and Analysis}

The experiments were performed on Google Colab Pro with 25 GB RAM and the codes were developed using python 3. We have used VGG16, InceptionV3, ResNet50, Xception, MobileNetV2, and EfficientNetB7 baseline unimodals to perform experiments. Adam optimizer with a learning rate of $1 \times 10^{-5}$ is employed to train the neural network models. We used F1-score, Precision, Recall, AUC, Accuracy metric for model evaluation. Different other libraries are used for the experimentation, such as OpenCV, Keras, sklearn, Scipy, Pandas, and face recognition. 








\begin{table*}[!h]
\centering
\caption{Comparison of the performance of the hybrid model on various datasets w.r.t the number of frames}
\resizebox{\linewidth}{!}{%
\begin{tabular}{|c|ccccccccc|}
\hline
\multicolumn{1}{|c|}{} &
  \multicolumn{3}{c|}{\textbf{DFDC Dataset}} &
  \multicolumn{3}{c|}{\textbf{FF++ Dataset}} &
  \multicolumn{3}{c|}{\textbf{CelebDF Dataset}} \\ \hline
 \multicolumn{1}{|c|}{\textbf{Frames}} & 
  \multicolumn{1}{c|}{\textbf{Accuracy}} &
  \multicolumn{1}{c|}{\textbf{Precision}} &
  \multicolumn{1}{l|}{\textbf{AUC Score}} &
  \multicolumn{1}{l|}{\textbf{Accuracy}} &
  \multicolumn{1}{l|}{\textbf{Precision}} &
  \multicolumn{1}{l|}{\textbf{AUC Score}} &
  \multicolumn{1}{l|}{\textbf{Accuracy}} &
  \multicolumn{1}{l|}{\textbf{Precision}} &
  \multicolumn{1}{l|}{\textbf{AUC Score}} \\ \hline
\textbf{10} &
  \multicolumn{1}{c|}{64.58\%} &
  \multicolumn{1}{c|}{64.39\%} &
  \multicolumn{1}{c|}{0.63} &
  \multicolumn{1}{c|}{74.87\%} &
  \multicolumn{1}{c|}{75.71\%} &
  \multicolumn{1}{c|}{0.75} &
  \multicolumn{1}{c|}{63.25\%} &
  \multicolumn{1}{c|}{66.67\%} &
  0.65 \\ \hline
\textbf{20} &
  \multicolumn{1}{c|}{64.27\%} &
  \multicolumn{1}{c|}{65.99\%} &
  \multicolumn{1}{c|}{0.64} &
  \multicolumn{1}{c|}{78.39\%} &
  \multicolumn{1}{c|}{78.80\%} &
  \multicolumn{1}{c|}{0.78} &
  \multicolumn{1}{c|}{67.09\%} &
  \multicolumn{1}{c|}{68.64\%} &
  0.68 \\ \hline
\textbf{30} &
  \multicolumn{1}{c|}{\textbf{66.26\%}} &
  \multicolumn{1}{c|}{67.11\%} &
  \multicolumn{1}{c|}{\textbf{0.66}} &
  \multicolumn{1}{c|}{83.17\%} &
  \multicolumn{1}{c|}{83.26\%} &
  \multicolumn{1}{c|}{0.83} &
  \multicolumn{1}{c|}{73.07\%} &
  \multicolumn{1}{c|}{72.94\%} &
  0.73 \\ \hline
\textbf{40} &
  \multicolumn{1}{c|}{64.12\%} &
  \multicolumn{1}{c|}{\textbf{69.04\%}} &
  \multicolumn{1}{c|}{0.63} &
  \multicolumn{1}{c|}{77.89\%} &
  \multicolumn{1}{c|}{82.31\%} &
  \multicolumn{1}{c|}{0.79} &
  \multicolumn{1}{c|}{73.08\%} &
  \multicolumn{1}{c|}{72.84\%} &
  0.73 \\ \hline
\textbf{50} &
  \multicolumn{1}{c|}{-} &
  \multicolumn{1}{c|}{-} &
  \multicolumn{1}{c|}{-} &
  \multicolumn{1}{c|}{86.68\%} &
  \multicolumn{1}{c|}{86.95\%} &
  \multicolumn{1}{c|}{0.87} &
  \multicolumn{1}{c|}{78.21\%} &
  \multicolumn{1}{c|}{78.19\%} &
  0.78 \\ \hline
\textbf{60} &
  \multicolumn{1}{c|}{-} &
  \multicolumn{1}{c|}{-} &
  \multicolumn{1}{c|}{-} &
  \multicolumn{1}{c|}{83.17\%} &
  \multicolumn{1}{c|}{86.32\%} &
  \multicolumn{1}{c|}{0.84} &
  \multicolumn{1}{c|}{72.65\%} &
  \multicolumn{1}{c|}{76.33\%} &
  0.73 \\ \hline
\textbf{70} &
  \multicolumn{1}{c|}{-} &
  \multicolumn{1}{c|}{-} &
  \multicolumn{1}{c|}{-} &
  \multicolumn{1}{c|}{\textbf{91.21\%}} &
  \multicolumn{1}{c|}{\textbf{91.20\%}} &
  \multicolumn{1}{c|}{\textbf{0.91}} &
  \multicolumn{1}{c|}{74.36\%} &
  \multicolumn{1}{c|}{76.34\%} &
  0.74 \\ \hline
\textbf{80} &
  \multicolumn{1}{c|}{-} &
  \multicolumn{1}{c|}{-} &
  \multicolumn{1}{c|}{-} &
  \multicolumn{1}{c|}{-} &
  \multicolumn{1}{c|}{-} &
  \multicolumn{1}{c|}{-} &
  \multicolumn{1}{c|}{76.07\%} &
  \multicolumn{1}{c|}{81.83\%} &
  0.76 \\ \hline
\textbf{90} &
  \multicolumn{1}{c|}{-} &
  \multicolumn{1}{c|}{-} &
  \multicolumn{1}{c|}{-} &
  \multicolumn{1}{c|}{-} &
  \multicolumn{1}{c|}{-} &
  \multicolumn{1}{c|}{-} &
  \multicolumn{1}{c|}{\textbf{79.49\%}} &
  \multicolumn{1}{c|}{\textbf{82.69\%}} &
  \textbf{0.79} \\ \hline
\textbf{100} &
  \multicolumn{1}{c|}{-} &
  \multicolumn{1}{c|}{-} &
  \multicolumn{1}{c|}{-} &
  \multicolumn{1}{c|}{-} &
  \multicolumn{1}{c|}{-} &
  \multicolumn{1}{c|}{-} &
  \multicolumn{1}{c|}{76.07\%} &
  \multicolumn{1}{c|}{79.92\%} &
  0.76 \\ \hline
\end{tabular}%
}
\label{tab:table5}
\end{table*}

\begin{figure*}[!h]
     \centering
     \begin{subfigure}{0.31\linewidth}
         \centering
         \includegraphics[width=\linewidth]{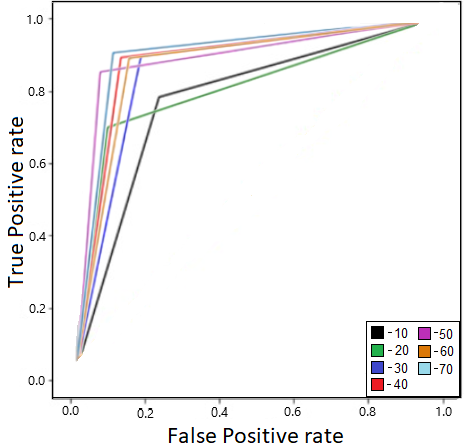}
         \caption{FaceForensics++}
         \label{fig:FF}
     \end{subfigure}%
     \hfill
     \begin{subfigure}{0.31\linewidth}
         \centering
         \includegraphics[width=\linewidth]{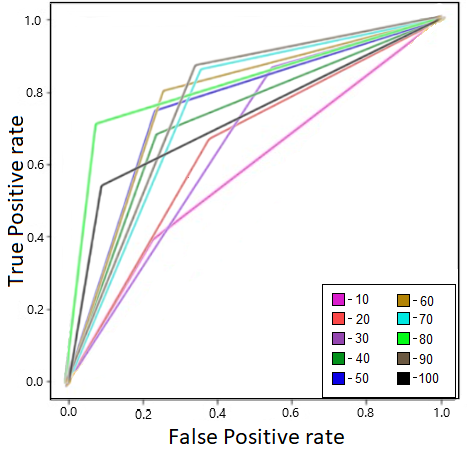}
         \caption{Celeb DF}
         \label{fig:CelebDF}
     \end{subfigure}%
     \hfill
     \begin{subfigure}{0.34\linewidth}
         \centering
         \includegraphics[width=\linewidth]{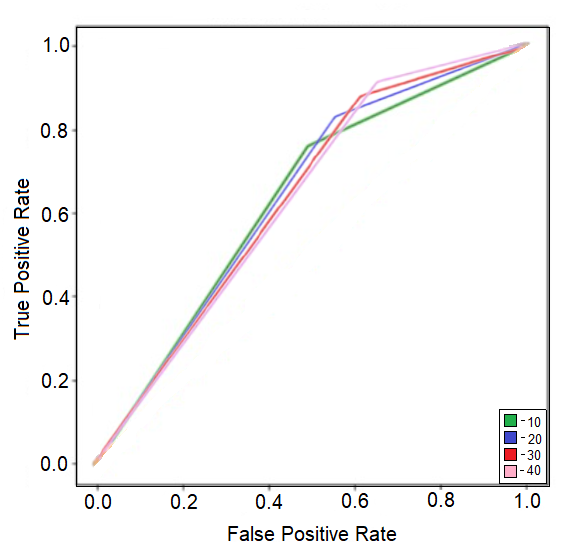}
         \caption{Deepfake Detection Challenge Dataset}
         \label{fig:DFDC}
     \end{subfigure}%
      \hfill
        \caption{AUC curves for the three datasets. Here the color indicates the corresponding number of frames as indicated in labels in respective plots.}
        \label{fig:three graphs}
\end{figure*}

We first experimented by varying the pre-trained CNN models as shown in table \ref{tab:table6}. As we can see, InceptionV3 is not  able  to  differentiate  well  as it gives out almost 0.5 value for the AUC curve. For the DFDC dataset, VGG16 is giving better performance with 20 Frames. EfficientNetB7 is giving better accuracy for Celeb-DF dataset with 50 frames and VGG16 is the second best for the same. For FF++, Resnet50 is giving higher accuracy with 30 frames and VGG16 is the third best performing for FF++. Taking all metrics into consideration ResNet 50 performs best for all the three datasets for the reason that it is faster to train and easier to use and deploy. Despite the fact that ResNet 50 is much deeper than VGG16 the model size is significantly smaller because of the utilization of global average pooling rather than fully-connected layers - this diminishes the model size down perhaps that might be the reason why it obtains better result than other models. However, all the models don't have much differences in  performance irrespective of the pre-trained architecture chosen.

Table \ref{tab:table4} provides the comparison result of the proposed model with various compositions of the models components: optical flow field (OF), CNN and RNN. From the results on experimentation on 40 frames, 70 frames and 100 frames of DFDC, FF++ and Celeb-DF datasets, it can be deduced that, the (OF+RNN) model is not able to differentiate well as all the datasets gives out 0.5 value for the AUC curve. This means the classifier is ineffective to properly distinguish between positive and negative class points, and hence for all available data points, it predicts a constant or random a class. The (OF+CNN) model performs better results than previous model and can distinguish between Real and Fake. For the Celeb DF and DFDC, (OF+CNN) model yields finer results than the hybrid (OF+CNN+RNN) model. On FF++ Dataset, the hybrid model outperforms both the other models.

Figures~\ref{fig:FF},~\ref{fig:CelebDF}, and ~\ref{fig:DFDC} illustrate the change in the AUC (Area Under The Curve) curve on the considered dataset by our proposed model (OF+RNN+CNN). AUC score of 0.5 means model cannot distinguish between real or fake. AUC score of 1 depicts that the model is able to perfectly classify the videos into real or fake. Thus an excellent model has an AUC score closer to 1. The AUC curves obtained for our hybrid model at varying number of frames have been merged into a single graph for each dataset, to showcase how the model reaches better performance with increasing number of frames. The different coloured lines represent curves at different number of frames. For FF++ as shown in Figure \ref{fig:FF}, the curves up to 70 frames have been considered. The light blue curve corresponds to an AUC score of 0.91, which is the highest the model has achieved. For CelebDF dataset as shown in Figure \ref{fig:CelebDF} the curves up to 100 frames have been plotted. It contains a total of 1168 videos, the least as compared to DFDC and FaceForensics++. For the DFDC dataset as shown in Figure \ref{fig:DFDC}, plotted curves up to 40 frames have been considered, as DFDC contains 3293 videos, the highest as compared to other two datasets used in this paper.

\begin{figure}[!h]
\centering
\includegraphics[width=\linewidth]{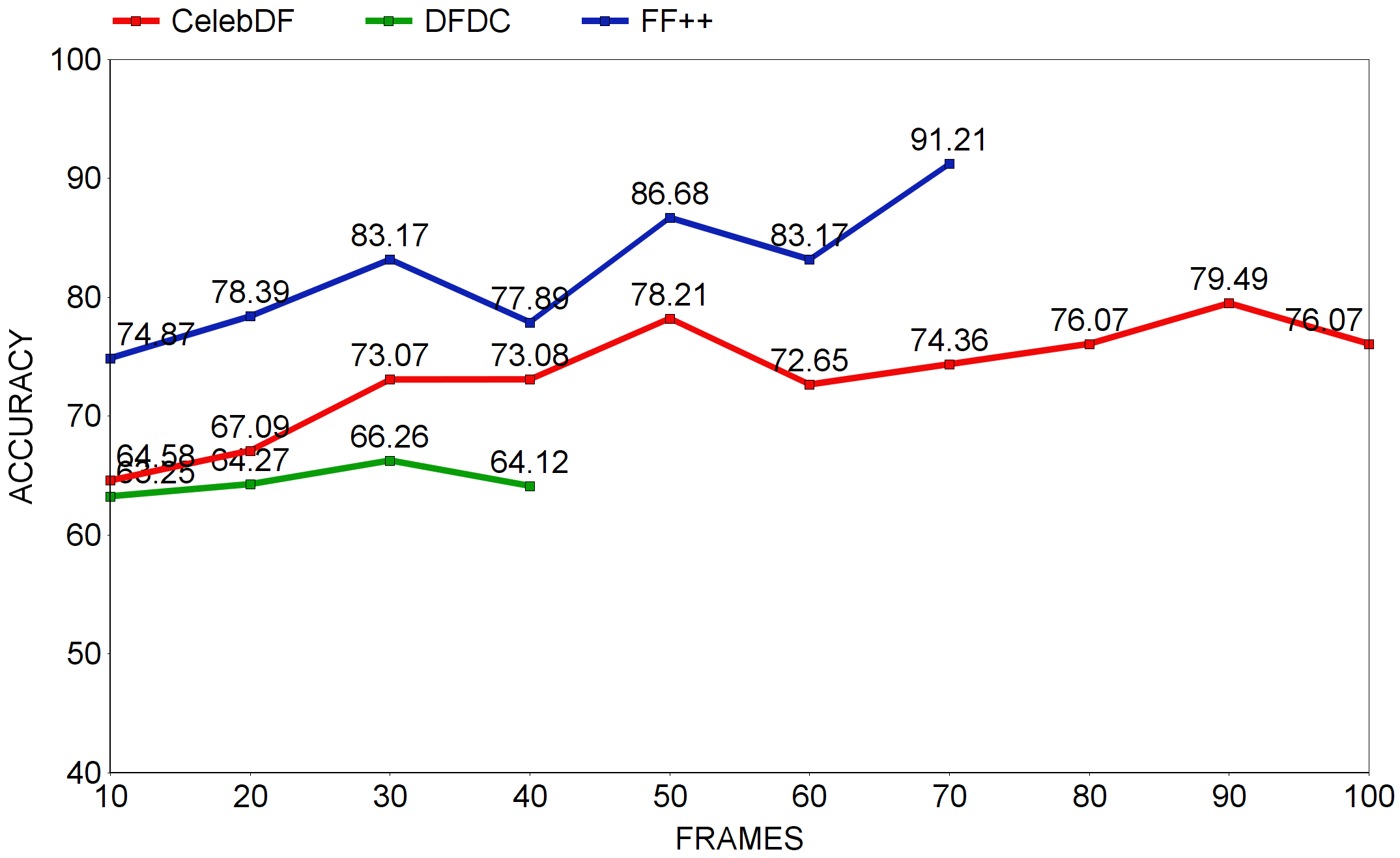}
\caption{Performance of the 3 datasets w.r.t the number of frame. Here the color indicates the corresponding dataset.}
\label{fig:figname1}
\centering
\end{figure}

Table \ref{tab:table5} provides the detailed comparison of the proposed model on varying the number of frames used per video for training and testing. Figure~\ref{fig:figname1} illustrates the observations from the table \ref{tab:table5} in a graphical manner. From the table it is observed that there is a general trend that the performance of the model improves as the number of frames increases. This is because with more frames per video the model can better detect the temporal variances for classification as real or fake. There have been certain deviations from this trend but overall performance has increased. FaceForensics++ was the best performing dataset and DFDC being the dataset with the least performance metrics. In this table only frames upto 40 have been considered for DFDC dataset and upto 70 frames for FaceForensics++ to reduce computational complexity. Most of the work in the area of Deepfake detection has used more frames than used in our work. There is no doubt that with the increase in the number of frames, the accuracy of our model would also increase to a great extent.

\begin{table}[!h]
\centering
\caption{Comparing proposed model with existing works.}
\label{tab:table3}
\resizebox{\linewidth}{!}{%
\begin{tabular}{|c|c|c|cc|c|c|}
\hline
Paper &
  Method &
  Model &
  \multicolumn{2}{c|}{Datasets} &
  Accuracy &
  AUC \\ \hline
\multirow{2}{*}{\begin{tabular}[c]{@{}c@{}}Irene \\ Amerini, \\ et. al. \cite{amerini2019deepfake}\end{tabular}} &
  \multirow{2}{*}{\begin{tabular}[c]{@{}c@{}}Optical \\ flow +\\  CNN\end{tabular}} &
  VGG16 &
  \multicolumn{2}{c|}{FF++} &
  81.61\% &
  - \\ \cline{3-7} 
 &
   &
  \begin{tabular}[c]{@{}c@{}}ResNet\\ 50\end{tabular} &
  \multicolumn{2}{c|}{FF++} &
  75.46\% &
  - \\ \hline
\multirow{5}{*}{\begin{tabular}[c]{@{}c@{}}Peng \\ Chen, \\ et. al.\\ \cite{chen2020fsspotter}\end{tabular}} &
  \multirow{5}{*}{\begin{tabular}[c]{@{}c@{}}CNN \& \\ LSTM\end{tabular}} &
  \multirow{5}{*}{VGG16} &
  \multicolumn{2}{c|}{FF++} &
  - &
  100 \\ \cline{4-7} 
 &
   &
   &
  \multicolumn{2}{c|}{Celeb-DF} &
  - &
  77.6 \\ \cline{4-7} 
 &
   &
   &
  \multicolumn{2}{c|}{UADFV} &
  - &
  91.1 \\ \cline{4-7} 
 &
   &
   &
  \multicolumn{1}{c|}{\multirow{2}{*}{\begin{tabular}[c]{@{}c@{}}Deepfake\\ TIMIT\end{tabular}}} &
  HQ &
  - &
  98.5 \\ \cline{5-7} 
 &
   &
   &
  \multicolumn{1}{c|}{} &
  LQ &
  - &
  99.5 \\ \hline
\multirow{4}{*}{\begin{tabular}[c]{@{}c@{}}D \\ Afchar,  \\ et  al.\\ \cite{afchar2018mesonet}\end{tabular}} &
  \multirow{4}{*}{CNN} &
  \multirow{2}{*}{Meso-4,} &
  \multicolumn{2}{c|}{\begin{tabular}[c]{@{}c@{}}FF++\\ (Face2face)\end{tabular}} &
  95\% &
  - \\ \cline{4-7} 
 &
   &
   &
  \multicolumn{2}{c|}{Deepfake} &
   &
  - \\ \cline{3-7} 
 &
   &
  \multirow{2}{*}{\begin{tabular}[c]{@{}c@{}}Meso-\\ Inception-4\end{tabular}} &
  \multicolumn{2}{c|}{\begin{tabular}[c]{@{}c@{}}FF++\\ (Face2face)\end{tabular}} &
  98\% &
  - \\ \cline{4-7} 
 &
   &
   &
  \multicolumn{2}{c|}{Deepfake} &
   &
  - \\ \hline
\multirow{6}{*}{\begin{tabular}[c]{@{}c@{}}Shivangi  \\ Aneja,  \\ et  al.\\ \cite{aneja2020generalized}\end{tabular}} &
  \multirow{6}{*}{\begin{tabular}[c]{@{}c@{}}CNN \& \\ LSTM\end{tabular}} &
  \multirow{6}{*}{Resnet18} &
  \multicolumn{2}{c|}{FF++} &
  92.23\% &
  - \\ \cline{4-7} 
 &
   &
   &
  \multicolumn{2}{c|}{Google DFD} &
  81.21\% &
  - \\ \cline{4-7} 
 &
   &
   &
  \multicolumn{2}{c|}{AIF} &
  60.79\% &
  - \\ \cline{4-7} 
 &
   &
   &
  \multicolumn{2}{c|}{Dessa} &
  74.28\% &
  - \\ \cline{4-7} 
 &
   &
   &
  \multicolumn{2}{c|}{Celeb-DF} &
  68.83\% &
  - \\ \cline{4-7} 
 &
   &
   &
  \multicolumn{2}{c|}{Combined} &
  75.47\% &
  - \\ \hline
\multirow{4}{*}{\begin{tabular}[c]{@{}c@{}}Pranjal \\ Ranjan, \\ et. al\\ \cite{ranjan2020improved}\end{tabular}} &
  \multirow{4}{*}{\begin{tabular}[c]{@{}c@{}}CNN + \\ LSTM\end{tabular}} &
  \multirow{4}{*}{\begin{tabular}[c]{@{}c@{}}Xception\\ Net\end{tabular}} &
  \multicolumn{2}{c|}{Celeb-DF} &
  83.49\% &
  - \\ \cline{4-7} 
 &
   &
   &
  \multicolumn{2}{c|}{DFDC} &
  78.13\% &
  - \\ \cline{4-7} 
 &
   &
   &
  \multicolumn{2}{c|}{DFD} &
  94.33\% &
  - \\ \cline{4-7} 
 &
   &
   &
  \multicolumn{2}{c|}{Combined} &
  79.62\% &
  - \\ \hline
\multirow{3}{*}{\begin{tabular}[c]{@{}c@{}}X Li, \\ et. al\\ \cite{li2020sharp}\end{tabular}} &
  \multirow{3}{*}{\begin{tabular}[c]{@{}c@{}}Multiple \\ Instance \\ Learning\end{tabular}} &
  \multirow{3}{*}{\begin{tabular}[c]{@{}c@{}}Xception\\ Net\end{tabular}} &
  \multicolumn{2}{c|}{Celeb-DF} &
  85.11\% &
  - \\ \cline{4-7} 
 &
   &
   &
  \multicolumn{2}{c|}{DFDC} &
  98.84\% &
  - \\ \cline{4-7} 
 &
   &
   &
  \multicolumn{2}{c|}{FFPMS} &
  90.71\% &
  - \\ \hline
\multirow{5}{*}{\begin{tabular}[c]{@{}c@{}}De \\ Lima, \\ et. al\\ \cite{de2020deepfake}\end{tabular}} &
  \multirow{5}{*}{\begin{tabular}[c]{@{}c@{}}Spatio-\\temporal \\ Convolutional \\ Networks\end{tabular}} &
  RCN &
  \multicolumn{2}{c|}{\multirow{5}{*}{Celeb-DF}} &
  76.25\% &
  74.87 \\ \cline{3-3} \cline{6-7} 
 &
   &
  R2Plus1D &
  \multicolumn{2}{c|}{} &
  98.07\% &
  99.43 \\ \cline{3-3} \cline{6-7} 
 &
   &
  I3D &
  \multicolumn{2}{c|}{} &
  92.28\% &
  97.59 \\ \cline{3-3} \cline{6-7} 
 &
   &
  R3D &
  \multicolumn{2}{c|}{} &
  98.26\% &
  99.73 \\ \cline{3-3} \cline{6-7} 
 &
   &
  MC3 &
  \multicolumn{2}{c|}{} &
  97.49\% &
  99.30 \\ \hline
\begin{tabular}[c]{@{}c@{}}SA \\ Khan,\\  et. al\\ \cite{khan2021adversarially} \end{tabular} &
  CNN &
  VGG16 &
  \multicolumn{2}{c|}{DFDC} &
  96.75\% &
  - \\ \hline
\multirow{3}{*}{\begin{tabular}[c]{@{}c@{}}Our \\ Work\end{tabular}} &
  \multirow{3}{*}{\begin{tabular}[c]{@{}c@{}}Optical Flow \\ + CNN \\ + LSTM\end{tabular}} &
  \multirow{3}{*}{VGG16} &
  \multicolumn{2}{c|}{FF++} &
  91.21\% &
  0.91 \\ \cline{4-7} 
 &
   &
   &
  \multicolumn{2}{c|}{Celeb-DF} &
  79.49\% &
  0.79 \\ \cline{4-7} 
 &
   &
   &
  \multicolumn{2}{c|}{DFDC} &
  66.26\% &
  0.66 \\ \hline
\end{tabular}%
}
\end{table}

Table \ref{tab:table3} provides the results of our proposed model with the baseline models in the literature. One study on the effect of optical flow and CNN training was also discussed in the work \cite{amerini2019deepfake}. Work by Peng Chen et. al. \cite{chen2020fsspotter} investigates the rich  spatial features with the help of spatial and temporal clues. They employ a two stage training strategy by learning temporal features and spatial inconsistencies separately. They have only used the AUC score percentage to present their results. Another work \cite{aneja2020generalized} have adopted the approach of Deep Distribution Transfer learning. Results are obtained with 4 datasets individually as well by combining all the datasets. Comparing their results on Celeb-DF dataset with the results we obtained, our model has achieved higher accuracy. Other works like \cite{afchar2018mesonet,ranjan2020improved,li2020sharp,de2020deepfake,khan2021adversarially} have presented better performance only based on accuracy. The reason for this can be accounted to the fact that our work was limited to a certain number of frames of the videos. 
In this paper we are using CNN along with the LSTM and then we are further improving it by combining optical flow features. Our model has been evaluated on 5 metrices namely accuracy, F1-score, Precision, Recall, and AUC, for Celeb-DF they are 79.80\%, 78.80\%, 82.49\%, 79.08\%, and 0.79, respectively, for FF++, they are 91.21\%, 91.20\%, 91.20\%, 91.21\%, and 0.91, respectively, and for DFDC, they are 66.26\%, 65.35\%, 67.11\%, 65.73\%, and 0.66, respectively. 
, whereas most of the work in this field evaluate their performance only on accuracy and AUC score. From this comparision we can find that our model is showing a comparable result even with a reduced frame number.

\section{Conclusion and Future Scope}

This work is based on the use of Optical Flow vectors with pre-trained CNN model, appended with LSTM layers to model the inconsistent motion of each pixel of the frames of videos, which can be evaluated to classify a video into fake or real. To reduce the computational constraints, the experiment was performed on a subset of frames as considering all the frames of the videos require higher computational power. However, from the experimentation it is observed that the model performed better with an increasing number of frames per video. Our work paves the way for many possible future works: firstly the model can be improved by training on huge set of the frames of the videos.  Secondly, more datasets can be incorporated for better performance so that the model can be trained to detect videos of all kinds of deepfake manipulation techniques. Further the comparable score of our proposed model with the reduced number of frames indicate the possible realization of early detection of the fake content. Thus, the application of optical flow field seems to be promising in this domain and can be further investigated on explainability of ultra-realistic deepfakes.




\bibliographystyle{IEEEtran}
\bibliography{reference}





\end{document}